%% file: MeT_arxiv.tex
\documentclass[10pt,twocolumn,letterpaper]{article}
\usepackage[pagenumbers]{cvpr}
\input{preamble}
\usepackage{marvosym}

\definecolor{cvprblue}{rgb}{0.21,0.49,0.74}
\usepackage[pagebackref,breaklinks,colorlinks,allcolors=cvprblue]{hyperref}

\title{Native Scalable Memory Models with Transformers}
\author{%
    Ziyang Xu\textsuperscript{1}\enspace
    Lucia Cascone\textsuperscript{2}\enspace
    Aniello Castiglione\textsuperscript{2}\enspace
    Michele Nappi\textsuperscript{2}\enspace
    Wenyu Liu\textsuperscript{1}\enspace
    Xinggang Wang\textsuperscript{1,}\thanks{Corresponding author: Xinggang Wang \textless\href{mailto:xgwang@hust.edu.cn}{xgwang@hust.edu.cn}\textgreater.}\\[0.35em]
    {\normalsize
        \textsuperscript{1}Huazhong University of Science and Technology\qquad
        \textsuperscript{2}University of Salerno}
}
\hypersetup{
    pdftitle={\thetitle},
    pdfauthor={Ziyang Xu, Lucia Cascone, Aniello Castiglione, Michele Nappi, Wenyu Liu, Xinggang Wang}
}

\begin{document}
\makeatletter
\begingroup
\renewcommand*{\@fnsymbol}[1]{\Letter}
\patchcmd{\@maketitle}{\vskip .375in}{\vskip 0pt}{}%
    {\PackageWarningNoLine{MeT_arxiv}{Could not adjust the space above the title}}
\patchcmd{\@maketitle}{\vspace*{24pt}}{\vspace*{14pt}}{}%
    {\PackageWarningNoLine{MeT_arxiv}{Could not adjust the title-to-author gap}}
\maketitle
\endgroup
\makeatother

\begin{abstract}
We introduce Memory Transformer (MeT), a novel method for native visual memory modeling that opens a new scaling dimension for vision Transformers. Its hierarchical memory maintains recurrent states updated from spatial features and returned through residual readouts, making persistent context an intrinsic part of the representation hierarchy. A separate hidden-state width controls memory capacity independently of backbone width and depth, while the retained state remains bounded as the input sequence grows. MeT reveals two complementary scaling trends: increasing memory capacity consistently improves accuracy at a fixed spatial backbone, and larger models deliver further gains at matched memory widths. Studies spanning three model sizes and fifteen capacity configurations demonstrate these gains with compact states and low computational overhead relative to same-size ViTs; per-frame latency remains nearly unchanged during memory scaling. On THUMOS14 and SSv2, MeT-Large outperforms state-of-the-art methods in accuracy, improving mAP and Top-1 accuracy over same-size ViTs by 6.69 and 26.91, respectively, with comparable parameter counts, GFLOPs, and GPU peak. Compared with TRecViT-Large, MeT-Large uses 92.7\% less persistent state and provides approximately $3.0\times$ faster causal framewise inference. MeT establishes native visual memory as an independently scalable architectural resource, opening a direct route to stronger visual models. \mbox{Project page at \url{https://hustvl.github.io/MeT}.}
\end{abstract}

\begin{figure}[!t]
\centering
\includegraphics[width=\linewidth]{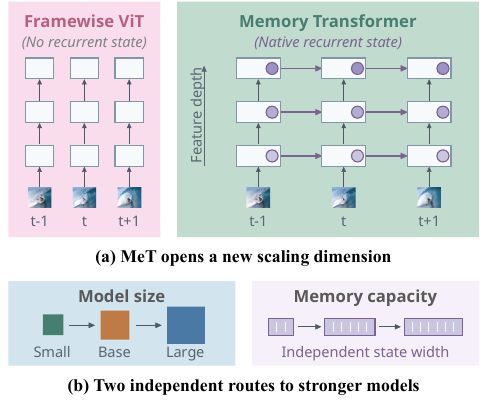}
\caption{\textbf{MeT makes native visual memory an independently scalable architectural resource.} Boxes denote spatial features; circles denote recurrent states carried forward within the same visual sequence. Spatial model size and memory capacity can grow independently, expanding the design space of visual Transformers.}
\label{fig:native_memory}
\end{figure}

\begin{figure*}[!t]
\centering
\includegraphics[width=\textwidth]{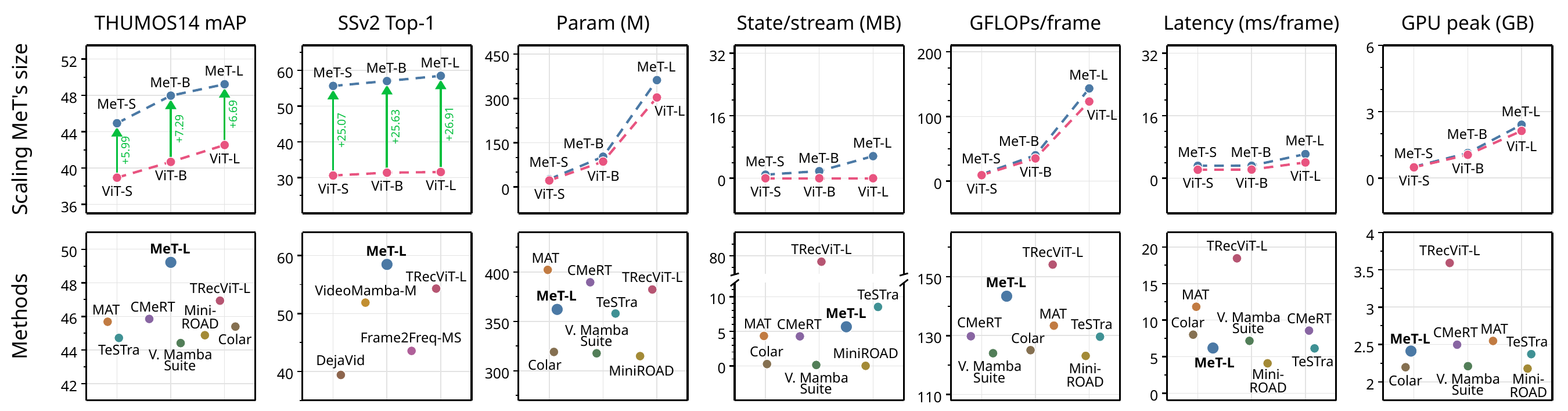}
\caption{\textbf{Scaling MeT's size improves accuracy with compact states and low computational overhead.} Fig.~\ref{fig:capacity} further shows consistent accuracy gains from scaling memory capacity. All models are trained on the same benchmark splits and evaluated with the same protocols. The five resource columns report THUMOS14 configurations; Tabs.~\ref{tab:01_thumos} and~\ref{tab:02_ssv2} provide detailed comparisons of accuracy, inference efficiency, and resource use on THUMOS14 and SSv2, respectively. State/stream measures persistent state and required cache per video. Latency is the mean time per frame during causal framewise inference for one video; GPU peak reports allocated VRAM for four 32-frame video clips processed together. \textit{Top-row:} Green arrows show accuracy gains over same-size ViTs. MeT-Small surpasses ViT-Large on both benchmarks with less than one-tenth of its parameters and per-frame FLOPs. \textit{Bottom-row:} All models use ViT-L backbones except VideoMamba-M, which retains its native Mamba backbone; V. Mamba Suite abbreviates Video Mamba Suite. MeT-L leads in accuracy while improving on TRecViT-L across all five resource and inference metrics.}
\label{fig:intro_comparison}
\end{figure*}

\section{Introduction}
\label{sec:introduction}

Scaling spatial width and depth has made vision Transformers (ViTs) increasingly capable~\cite{dosovitskiy2021vit,zhai2022scalingvit}. When observations arrive over time, retained states connect past information to current feature computation~\cite{yang2022recurring,zoran2026rvm}. We introduce Memory Transformer (MeT), a family of ViTs whose native hierarchical memory has a recurrent-state width independent of spatial backbone width and depth. MeT establishes two complementary scaling trends: wider recurrent states and larger MeT models each improve accuracy.

Existing designs differ in how temporal context enters current computation. Spatiotemporal attention directly aggregates features across frames~\cite{bertasius2021timesformer,liu2022videoswin,fan2021mvit}. Activation caching reuses past representations as attention context, with capacity adjusted through history length and compression~\cite{dai2019transformerxl,wu2022memvit,ji2022lamemo}. Memory-token and slot recurrence propagates compact summaries between segments, with capacity set by the retained vector count~\cite{bulatov2022recurrentmemory,wu2022memformer,hutchins2022blockrecurrent}, while recurrent visual states evolve as frames arrive and shape current feature computation~\cite{yang2022recurring,patraucean2026trecvit,zoran2026rvm}. Hierarchical retention and retrieval further organize context across functional roles and timescales~\cite{he2025hmt}. We ask how persistent visual states can be integrated into spatial feature learning so that memory capacity becomes an effective scaling axis.

MeT addresses this question through memory stages at multiple spatial feature depths, each maintaining distinct token-level GRU states. Each stage updates its states from the current spatial features and returns memory readouts through residual connections. The recurrent hidden width $H$ controls memory capacity without widening or deepening the spatial backbone. Updates use only the current input and prior state, enabling causal framewise inference. The retained state remains bounded as the sequence grows. Fig.~\ref{fig:native_memory} contrasts framewise processing with MeT and shows its independent memory-capacity and model-size axes.

We substantiate these scaling trends across three model sizes and fifteen capacity configurations on THUMOS14~\cite{jiang2014thumos} and SSv2~\cite{goyal2017something}. At fixed spatial backbones and memory layouts, wider recurrent states yield consistent accuracy gains; larger MeT models provide further gains, including at matched memory widths. Fig.~\ref{fig:intro_comparison} shows that MeT-Small surpasses ViT-Large on both benchmarks using less than one-tenth of its parameters and per-frame FLOPs. MeT-Base surpasses TRecViT-Large on both benchmarks with approximately 27\% of its parameters and 26\% of its per-frame FLOPs. Fig.~\ref{fig:capacity} shows that memory-capacity scaling improves accuracy with nearly unchanged per-frame latency within each model size. MeT makes recurrent-state capacity an efficient route to higher visual accuracy.

Our contributions are as follows:
\begin{itemize}
\item We introduce MeT, a novel method for native hierarchical memory modeling in vision Transformers, with independently scalable recurrent-state capacity. Token-level recurrence is integrated into the spatial feature hierarchy through residual memory modules, making persistent context an intrinsic part of the architecture while maintaining bounded state for each input sequence.
\item We establish two complementary scaling trends in memory capacity and model size. Studies spanning three model sizes and fifteen capacity configurations show consistent accuracy gains from increasing memory width at fixed backbone architecture, with further gains from enlarging the model family, including comparisons at matched memory widths.
\item We characterize the mechanisms, token inputs, insertion depths, and layouts that support this scaling behavior, and validate the resulting architecture on THUMOS14 and SSv2. MeT-Large outperforms state-of-the-art methods in accuracy while using 92.7\% less persistent state and providing approximately $3.0\times$ faster causal framewise inference than TRecViT-Large.
\end{itemize}

\begin{figure*}[!t]
\centering
\includegraphics[width=\textwidth]{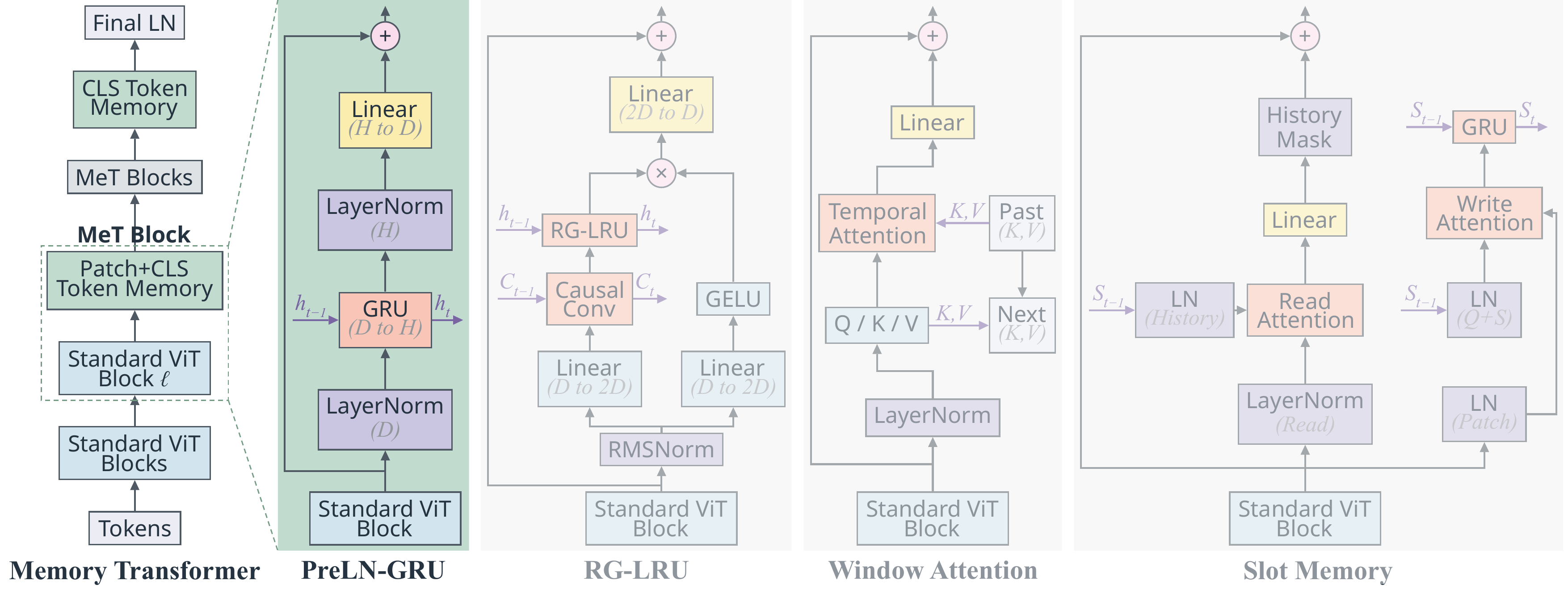}
\caption{\FigureFourCaption}
\label{fig:architecture}
\end{figure*}

\section{Related Work}
\label{sec:related_work}

\paragraph{Temporal Visual Representations.} Spatiotemporal attention extends spatial representations through cross-frame interactions~\cite{bertasius2021timesformer,arnab2021vivit,liu2022videoswin,li2022mvitv2}. Other clip-level designs encode temporal structure through bidirectional state-space processing~\cite{li2024videomamba,lu2025videomambapro}, frequency-selective adapters~\cite{ponbagavathi2026frame2freq}, or learned temporal matching~\cite{ho2025dejavid}. These mechanisms model temporal relations for clip-level prediction. MeT instead carries temporal information between frames through a bounded causal state. Prediction uses the current input and retained context without joint access to the full clip.

\paragraph{Memory for Streaming Recognition.} Streaming recognition often builds temporal memory over encoded visual features~\cite{xu2021lstr,wang2021oadtr}. Long- and short-range context can be combined through memory refinement and anticipation~\cite{pang2025cmert,wang2023memory}, summarized through temporal smoothing~\cite{zhao2022testra}, or complemented with exemplar retrieval~\cite{yang2022colar}. Recurrent aggregation and state-space modules also process these feature streams~\cite{an2023miniroad,chen2026videomambasuite,xu2019trn}. In these feature-stream settings, memory primarily organizes historical evidence for the task readout. MeT instead couples memory updates to the formation of spatial features at multiple depths, making persistent context part of the visual backbone itself.

\paragraph{Persistent States in Transformers.} Transformer memory differs in both its state representation and its interaction with feature computation~\cite{omidi2025memoryreview}. Activation caches reuse past features, optionally compressing older context~\cite{dai2019transformerxl,rae2020compressive,wu2022memvit,wang2023longmem}. Dedicated memory vectors participate in attention~\cite{burtsev2020memorytransformer}, while token- and slot-based recurrence carries summaries across segments~\cite{wu2022memformer,bulatov2022recurrentmemory}. Writable window memories~\cite{stangeland2026pmt} and layer-specific associative matrices~\cite{zhang2026metis,li2026memt,schlag2021fastweight} also participate in Transformer computation, alongside recurrent visual states~\cite{yang2022recurring,patraucean2026trecvit,zoran2026rvm}. MeT organizes its dynamic state by visual token position and spatial depth, with recurrent width separated from spatial feature width. Projected residual readouts return temporal information to the spatial tokens while preserving the backbone's feature interface.

\paragraph{Memory Hierarchies.} Hierarchical memory can distinguish functional roles and retention timescales~\cite{he2025hmt,mujawar2026hierarchicalhebbian}, or organize cached context across feature depths~\cite{wu2022memvit,zhu2026moda}. A shared functional hierarchy can serve several feature depths~\cite{mujawar2026hierarchicalhebbian}, whereas depth-specific caches retain states at their corresponding visual stages~\cite{wu2022memvit}. MeT's hierarchy consists of independently updated token states at distinct spatial depths. Each stage supplies its own temporal context as spatial features are formed.

\paragraph{Memory Capacity and Scaling.} Memory capacity has been varied through cache length and compression~\cite{wu2022memvit,rae2020compressive,wu2022memorizing,martins2022infiniteformer}, memory-token or slot counts~\cite{burtsev2020memorytransformer,wu2022memformer,bulatov2022recurrentmemory}, and the number of storage experts in associative memory~\cite{li2026memt}. Learned lookup capacity offers another scaling resource~\cite{berges2025memorylayers,choi2026vlumen}, distinct from the dynamic state carried by an input sequence. MeT fixes the spatial backbone architecture and memory layout when varying recurrent hidden width, separating dynamic visual-state capacity from spatial model capacity. By studying this axis together with model-family scaling, MeT establishes dynamic visual memory as an architectural resource that can be scaled in its own right.

\section{Memory Transformer}
\label{sec:method}

MeT interleaves ViT blocks with native recurrent memory whose capacity is controlled independently of spatial backbone width and depth. Each frame is represented by one CLS token and $n$ patch tokens, processed by $N$ spatial blocks of width $D$. Memory stages follow spatial blocks at multiple depths and return residual readouts to the token features. As Fig.~\ref{fig:architecture} illustrates, each MeT Block combines a Standard ViT Block with Token Memory while preserving the spatial feature width.

\begin{figure}[t]
\centering
\includegraphics[width=\columnwidth]{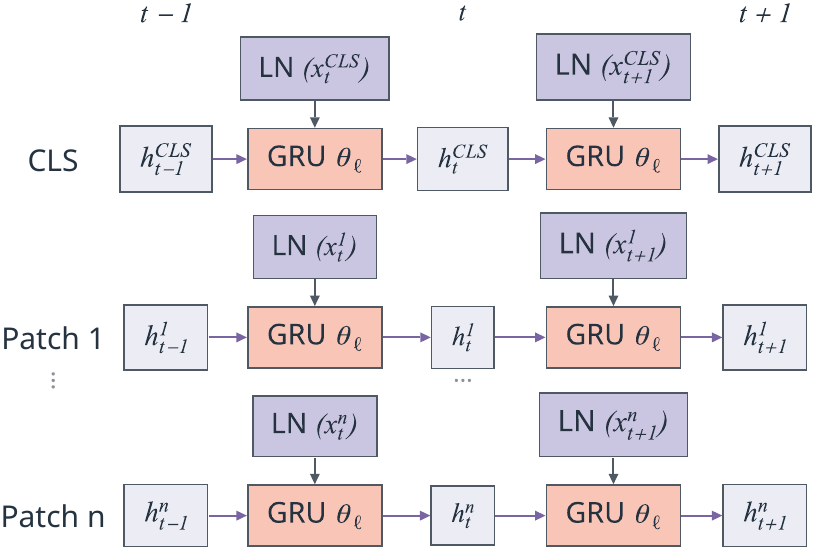}
\caption{\FigureFiveCaption}
\label{fig:token_states}
\end{figure}

\subsection{Native Memory Blocks}
\label{sec:token_recurrence}

\paragraph{Memory Block Designs.} We construct four memory-block designs to study how historical evidence should shape spatial computation, contrasting gated recurrence, direct access to recent features, and shared summaries across tokens. Fig.~\ref{fig:architecture} shows how each design follows a spatial block and preserves its feature width through a residual readout. For frame $t$, spatial block $\ell$, and token position $p\in\{\mathit{CLS},1,\ldots,n\}$, let $\mathbf{x}_{t}^{(\ell,p)}\in\mathbb{R}^{D}$ denote the current token feature.

\subparagraph{\normalfont\itshape RG-LRU.} To combine local temporal mixing with gated retention, we project RMS-normalized token features into two branches of width $2D$. One branch applies causal convolution to the current and three preceding projected inputs, then gated linear recurrence~\cite{de2024griffin}; the other provides a GELU-activated output gate. Their elementwise product is projected back to $D$ and added to the input. Each token retains its hidden state and the three latest projected inputs.

\begin{table}[t]
\input{tables/table_03_mechanism.tex}
\vspace{-12pt}
\end{table}

\begin{table}[t]
\input{tables/table_04_content_position.tex}
\end{table}

\begin{table*}[t]
\input{tables/table_05_layout.tex}
\end{table*}

\begin{table}[t]
\input{tables/table_06_family.tex}
\end{table}

\subparagraph{\normalfont\itshape Window Attention.} For direct access to recent evidence, we form queries, keys, and values from each token's normalized spatial feature. Temporal attention reads the current key and value together with up to eight preceding pairs at the same token position, and projects the result into a residual update. The current pair then enters the bounded cache, replacing the oldest entry when full. The cache stores keys and values projected before temporal attention.

\subparagraph{\normalfont\itshape Slot Memory.} To share historical summaries across token positions, we maintain eight $D$-dimensional slots. Normalized patch and CLS tokens query the normalized preceding slots through read attention, producing a residual update. For writing, learned slot queries are added to the preceding slots and normalized to attend to the original current patch features. A shared GRU updates each slot from the observation and prior state. Updated slots are read at the next frame; the first frame receives no slot readout.

\subparagraph{\normalfont\itshape PreLN-GRU.} To update token histories through nonlinear gating, we maintain a recurrent state $\mathbf{h}_{t}^{(\ell,p)}\in\mathbb{R}^{H}$ of hidden width $H$ for each token. The GRU~\cite{cho2014gru} uses parameters $\theta_\ell$; $\operatorname{LN}_{x}^{\ell}$ and $\operatorname{LN}_{h}^{\ell}$ denote input and hidden-state LayerNorm, respectively. An affine readout with matrix $\mathbf{W}_{o}^{\ell}\in\mathbb{R}^{D\times H}$ and bias $\mathbf{b}_{o}^{\ell}\in\mathbb{R}^{D}$ returns the normalized state to the spatial feature, yielding the updated token $\widetilde{\mathbf{x}}_{t}^{(\ell,p)}$. The raw hidden state is retained for the next frame before readout normalization. Zero-initializing $\mathbf{W}_{o}^{\ell}$ and $\mathbf{b}_{o}^{\ell}$ preserves the initial spatial feature computation.
\begin{align}
\mathbf{h}_{t}^{(\ell,p)} &= \operatorname{GRU}_{\theta_\ell}\!\left(\operatorname{LN}_{x}^{\ell}(\mathbf{x}_{t}^{(\ell,p)}),\mathbf{h}_{t-1}^{(\ell,p)}\right), \label{eq:state_update}\\
\widetilde{\mathbf{x}}_{t}^{(\ell,p)} &= \mathbf{x}_{t}^{(\ell,p)} + \mathbf{W}_{o}^{\ell}\operatorname{LN}_{h}^{\ell}(\mathbf{h}_{t}^{(\ell,p)})+\mathbf{b}_{o}^{\ell}. \label{eq:memory_readout}
\end{align}

Tab.~\ref{tab:03_mechanism} compares the four complete memory designs at the common mid-depth of ViT-Small. PreLN-GRU achieves the highest accuracy on both benchmarks and is used throughout MeT.

\paragraph{Parameter and State Organization.} Fig.~\ref{fig:token_states} unfolds this recurrence across token positions and time. At intermediate memory stages, the GRU, normalization, and readout parameters are shared across positions and time steps; CLS and each patch position retain their own state vectors. Different depths use distinct modules and states. Spatial blocks mix token features before the updates and mix the resulting features at subsequent depths, allowing position-specific histories to influence other tokens through spatial computation. States are initialized to zero for each new sequence.

\subsection{Hierarchical Memory Organization}
\label{sec:hierarchical_memory}

\paragraph{Memory Inputs and Depth.} The memory input determines which spatial evidence can be retained. Tab.~\ref{tab:04_content_position} studies patch tokens, CLS, and their combination, together with single-stage insertion depth. Retaining patch-token histories improves both benchmarks over CLS-only memory, and joint patch-and-CLS inputs achieve the highest accuracy. Moving the joint stage from block 6 to block 9 or 12 improves THUMOS14 mAP but reduces SSv2 Top-1 accuracy. These findings motivate joint patch-and-CLS memory at intermediate depths, where subsequent spatial mixing integrates memory readouts.

\paragraph{Hierarchical Placement.} Memory serves complementary roles across feature depths: intermediate patch-and-CLS histories shape subsequent spatial computation, while final-stage CLS memory refines the representation used for prediction. Fig.~\ref{fig:architecture} realizes this division through distinct memory stages at spaced depths. Tabs.~\ref{tab:04_content_position} and~\ref{tab:05_layout} show that distributing these stages after blocks 3, 6, 9, and 12 in MeT-Small improves both benchmarks over single-stage memory. This four-stage layout achieves the highest THUMOS14 mAP, retaining only 27.3\% of the state of all-block memory with strong SSv2 performance. We adopt this layout for MeT-Small and MeT-Base.

\begin{figure*}[!t]
\centering
\includegraphics[width=\textwidth]{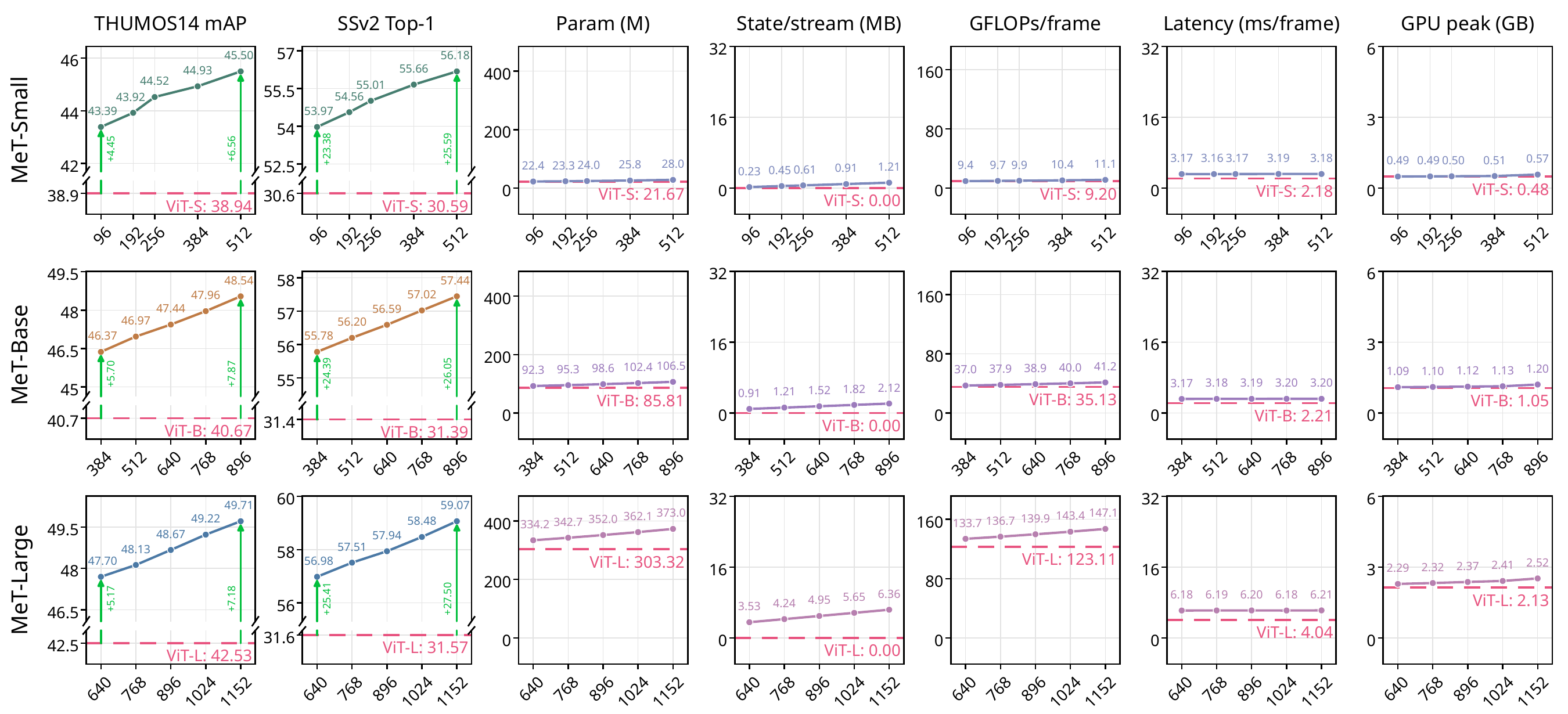}
\caption{\metcapacitycaption}
\label{fig:capacity}
\end{figure*}

\begin{samepage}
\subsection{Memory Capacity and Model Scaling}
\label{sec:memory_scaling}

\paragraph{Independent Memory Capacity.} The recurrent width $H$ controls memory capacity independently of spatial width $D$ and depth $N$: the GRU maps normalized inputs into $H$-dimensional states, and the residual readout maps them back to $D$. Memory-capacity scaling varies $H$ within the memory modules while fixing spatial width $D$, depth $N$, and the memory layout. For stage set $\mathcal{L}$, let $P_\ell$ denote the number of tokens updated at stage $\ell$, with $P_\ell=n+1$ at intermediate stages and $P_N=1$ at the final stage. The retained recurrent state does not grow with the number of processed frames. Counting FP32 state elements (4 bytes each) across memory stages gives:
\begin{equation}
\text{Recurrent state size}=H\sum_{\ell\in\mathcal{L}}P_\ell.
\label{eq:state_size}
\end{equation}
\end{samepage}

\paragraph{Model Configurations.} Tab.~\ref{tab:06_family} specifies the default MeT models. We match $H$ to ViT's hidden width and retain its spatial depth $N$. Keeping the same three-block memory spacing across these inherited depths yields four stages in MeT-Small and MeT-Base ($N=12$), and eight in MeT-Large ($N=24$). The default width alignment leaves $H$ independently adjustable within each model size. Increasing $H$ at a fixed backbone and layout scales memory capacity; scaling MeT's size enlarges the spatial backbone and its corresponding memory hierarchy. Fig.~\ref{fig:capacity} examines the accuracy and resource trends along these complementary architectural scaling axes; Sec.~\ref{sec:scaling_results} provides a detailed analysis.

\FloatBarrier

\section{Experiments}
\label{sec:experiments}

\subsection{Implementation Details}
\label{sec:implementation}

\begin{table*}[!t]
\input{tables/table_01_thumos.tex}
\end{table*}

\begin{table*}[!t]
\input{tables/table_02_ssv2.tex}
\end{table*}

\paragraph{Benchmarks and Inputs.} We evaluate causal online action recognition at the frame level on THUMOS14~\cite{jiang2014thumos} and clip-level video classification on SSv2~\cite{goyal2017something}. Following established online action recognition protocols~\cite{xu2021lstr,an2023miniroad}, we train on all 200 THUMOS14 validation videos and evaluate on the 213 test videos. We report 20-class framewise mAP at 4Hz using current and past observations, masking ambiguous intervals in the loss and mAP computation~\cite{xu2021lstr}. The 174-class SSv2 dataset uses its official split of 168,913 training and 24,777 validation videos~\cite{ssv2official}. Following segment-based sampling~\cite{wang2016tsn}, we draw one frame from each of 32 ordered temporal segments and evaluate a single temporal view with a center crop. Both benchmarks use $224\times224$ RGB inputs; SSv2 reports Top-1 and Top-5 accuracy, following video Transformer practice~\cite{liu2022videoswin,patraucean2026trecvit}.

\paragraph{MeT Training.} We initialize the spatial backbone with official ImageNet-1K-pretrained ViT weights~\cite{dosovitskiy2021vit,russakovsky2015imagenet} and train MeT using AdamW~\cite{loshchilov2019adamw} with weight decay 0.05. Training spans 40 epochs on THUMOS14 and 35 on SSv2~\cite{patraucean2026trecvit}. To learn from continuous videos while carrying states across updates, THUMOS14 training uses eight parallel streams with 32-frame truncated backpropagation through time~\cite{williams1990efficient}. SSv2 uses an effective batch of 64 clips, backpropagating through all 32 frames. Following the lower-backbone-rate fine-tuning practice~\cite{liu2022videoswin}, learning rates for the backbone, memory modules, and classifier are $(10^{-5},10^{-4},10^{-3})$ on THUMOS14 and $(5\times10^{-5},10^{-4},5\times10^{-4})$ on SSv2, respectively. Linear warmup lasts two epochs on THUMOS14 and five on SSv2, followed by cosine decay~\cite{patraucean2026trecvit}.

\paragraph{Comparators and Readouts.} To compare temporal modeling under common task conditions, all comparison methods receive target-task training on the same benchmark splits and follow the same evaluation protocols. Across both benchmarks, we combine TRecViT's recurrent blocks~\cite{patraucean2026trecvit} with the ViT-Large spatial backbone used by MeT-Large and ViT-Large to compare temporal modeling at a fixed spatial architecture. On THUMOS14, they apply a linear classifier to each frame's CLS features; feature-stream comparators~\cite{pang2025cmert,an2023miniroad,wang2023memory,zhao2022testra,yang2022colar,chen2026videomambasuite} train their temporal modules on frozen features from target-task-trained ViT-Large encoders. On SSv2, MeT, ViT, and TRecViT apply a linear classifier to time-averaged CLS features~\cite{wang2016tsn} to keep temporal pooling and classifier structure consistent. For the clip-based SSv2 comparators, full-clip spectral processing~\cite{ponbagavathi2026frame2freq}, sequence alignment~\cite{ho2025dejavid}, and bidirectional encoding~\cite{li2024videomamba} are integral to prediction; we retain these computations and their associated readouts to preserve each method's temporal mechanism. Specifically, Frame2Freq-MS and DejaVid train over frozen, target-task-trained ViT-Large encoders: the former learns spectral adapters and a classifier with the same CLS averaging~\cite{ponbagavathi2026frame2freq}; the latter classifies sequences of frame features and class logits through DTW distances to training-set class centroids~\cite{ho2025dejavid}. VideoMamba-M fine-tunes its native bidirectional backbone from ImageNet-1K weights and reads the jointly encoded clip CLS token~\cite{li2024videomamba}.

\vspace{-6pt}
\paragraph{Evaluation and Efficiency.} Efficiency metrics are measured on one NVIDIA RTX PRO 5000 Blackwell GPU. As defined in Fig.~\ref{fig:intro_comparison}, State/stream counts per-video states and required caches, including temporal-averaging statistics. States are reset between videos. Parameters and GFLOPs cover the visual encoder and all inference components. Latency reports the mean time per frame for causal framewise inference on one video. GPU peak reports allocated VRAM for four 32-frame clips processed together.

\begin{figure*}[!t]
\centering
\includegraphics[width=\textwidth]{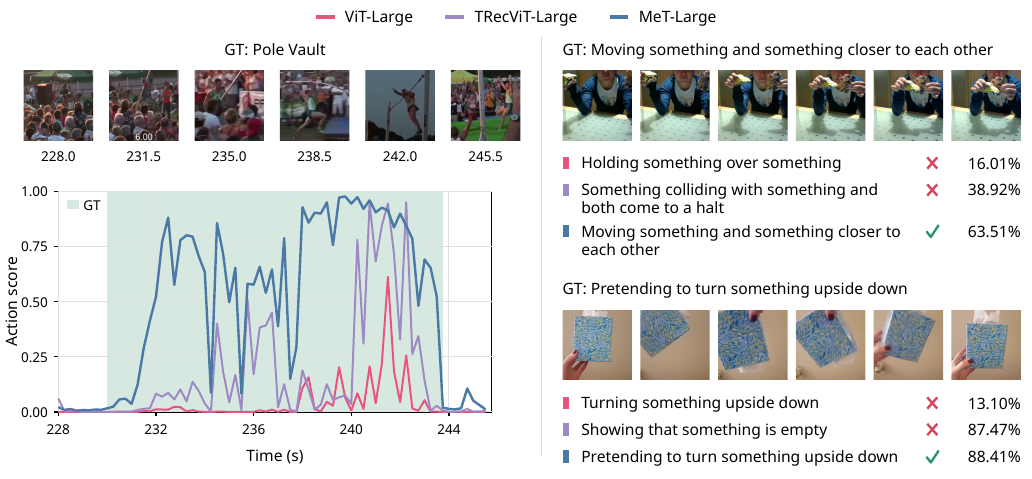}
\caption{\QualitativeCaption}
\label{fig:qualitative}
\end{figure*}

\subsection{Scaling Memory and Model Size}
\label{sec:scaling_results}

\paragraph{Scaling Memory Capacity.} Fig.~\ref{fig:capacity} reveals a consistent memory-scaling trend: increasing $H$ improves accuracy at every capacity for all three model sizes on both benchmarks. Each row fixes the spatial backbone architecture and memory layout. From the smallest to largest $H$, gains range from 2.01 to 2.17 mAP on THUMOS14 and 1.66 to 2.21 Top-1 accuracy on SSv2. These accuracy gains come at nearly constant framewise latency: the measured range within each model size is only 0.03ms per frame. Tab.~\ref{tab:05_layout} and Fig.~\ref{fig:capacity} separate memory placement from capacity: denser layouts do not consistently improve accuracy, whereas wider recurrent states at fixed layouts do.

\paragraph{Scaling Model Size.} Scaling the MeT family provides a complementary gain. Default MeT-Small, MeT-Base, and MeT-Large reach 44.93, 47.96, and 49.22 mAP on THUMOS14, and 55.66, 57.02, and 58.48 Top-1 accuracy on SSv2. Larger models also improve accuracy at overlapping memory widths in Fig.~\ref{fig:capacity}, separating this benefit from simply widening the recurrent states. Fig.~\ref{fig:intro_comparison} further shows that MeT-Small surpasses ViT-Large on both benchmarks with less than one-tenth of its parameters and per-frame FLOPs.

\subsection{Accuracy and Inference Efficiency}
\label{sec:comparison_results}

\paragraph{Causal Action Recognition.} Tab.~\ref{tab:01_thumos} compares accuracy and inference costs under the causal framewise protocol. MeT-Base already surpasses every comparator, and MeT-Large extends the lead to 49.22 mAP. Against TRecViT-Large, MeT-Large gains 2.28 mAP while retaining 92.7\% less state and providing approximately $3.0\times$ faster causal framewise inference. MeT-Large also reduces parameter count, FLOPs, and GPU peak.

\paragraph{Video Classification.} Tab.~\ref{tab:02_ssv2} shows that MeT-Small exceeds every non-MeT comparator. MeT-Large reaches 58.48 Top-1 and 84.77 Top-5 accuracy, improving over TRecViT-Large by 4.17 and 6.85, respectively, with lower costs across all five resource and inference metrics. MeT therefore preserves its advantage when temporal evidence is integrated into a complete-sequence decision.

\Needspace{6\baselineskip}
\subsection{Qualitative Results}
\label{sec:qualitative_results}

Fig.~\ref{fig:qualitative} connects these quantitative gains to predictions on real visual sequences. On THUMOS14, MeT-Large responds more strongly through the action interval while separating it from the surrounding background. On SSv2, it distinguishes approaching objects from collision, and a pretended action from its actual execution. These cases illustrate how evolving visual evidence supports discriminative judgments at both frame and sequence levels. Additional examples are provided in the \hyperref[sec:supp:qualitative]{supplementary material}.

\FloatBarrier

\Needspace{13\baselineskip}
\section{Conclusion}
\label{sec:conclusion}

We introduced MeT, a novel method for native hierarchical memory modeling that decouples memory capacity from spatial backbone width and depth. Recurrent token states and residual readouts make persistent context intrinsic to the spatial representation hierarchy. Scaling memory capacity and model size consistently improves accuracy, while memory scaling preserves nearly constant per-frame latency within each model size. On THUMOS14 and SSv2, MeT combines leading accuracy with compact states and efficient inference. MeT establishes persistent visual state as a scalable architectural resource, expanding the design space of ViTs beyond spatial model size.

\section*{LLM Usage Statement}

In this work, LLMs were used to assist with debugging and to check human-written text for typographical and grammatical errors.

{
    \small
    \bibliographystyle{ieeenat_fullname}
    \bibliography{main}
}

\appendix
\renewcommand{\thesection}{S\arabic{section}}

\twocolumn[{
    \pdfbookmark[0]{Supplementary Materials of MeT}{supplement}
    {\centering\Large\bfseries Supplementary Materials of MeT\par}
    \vspace{1em}
    {\input{tables/table_s01_size_comparison}}
    \vspace{1em}
}]

\section{Overview}
\label{sec:supp:overview}

\begin{itemize}
    \item \textbf{\hyperref[sec:supp:trecvit]{Sec.~\ref*{sec:supp:trecvit}\enspace Comparison with ViT and TRecViT.}} We compare the accuracy and inference efficiency of MeT, ViT, and TRecViT at matched backbone sizes.
    \item \textbf{\hyperref[sec:supp:qualitative]{Sec.~\ref*{sec:supp:qualitative}\enspace Additional Qualitative Results.}} We examine when models respond on THUMOS14 and how they distinguish similar actions on SSv2.
\end{itemize}

\section{Comparison with ViT and TRecViT}
\label{sec:supp:trecvit}

We compare ViT, TRecViT, and MeT at three matched backbone sizes in Tab.~\ref{tab:supp:size_comparison}. At each size, we initialize the three backbones from the same ImageNet-1K ViT weights and follow the task protocols in the main paper.

On SSv2, framewise ViT gains 0.98 Top-1 points from Small to Large. MeT-Small raises Top-1 by 25.07 points over ViT-Small. TRecViT applies tokenwise recurrence before each spatial block, and MeT inserts memory after every third block. At matched backbone sizes, MeT improves over TRecViT by 2.01--2.28 mAP on THUMOS14 and 3.55--4.17 Top-1 points on SSv2. On THUMOS14, MeT performs causal framewise inference approximately $2.8$--$3.0\times$ as fast as TRecViT, with fewer FLOPs per frame and less state per stream.

\newpage
\section{Additional Qualitative Results}
\label{sec:supp:qualitative}

\begin{figure}[!p]
    \centering
    \includegraphics[width=\textwidth]{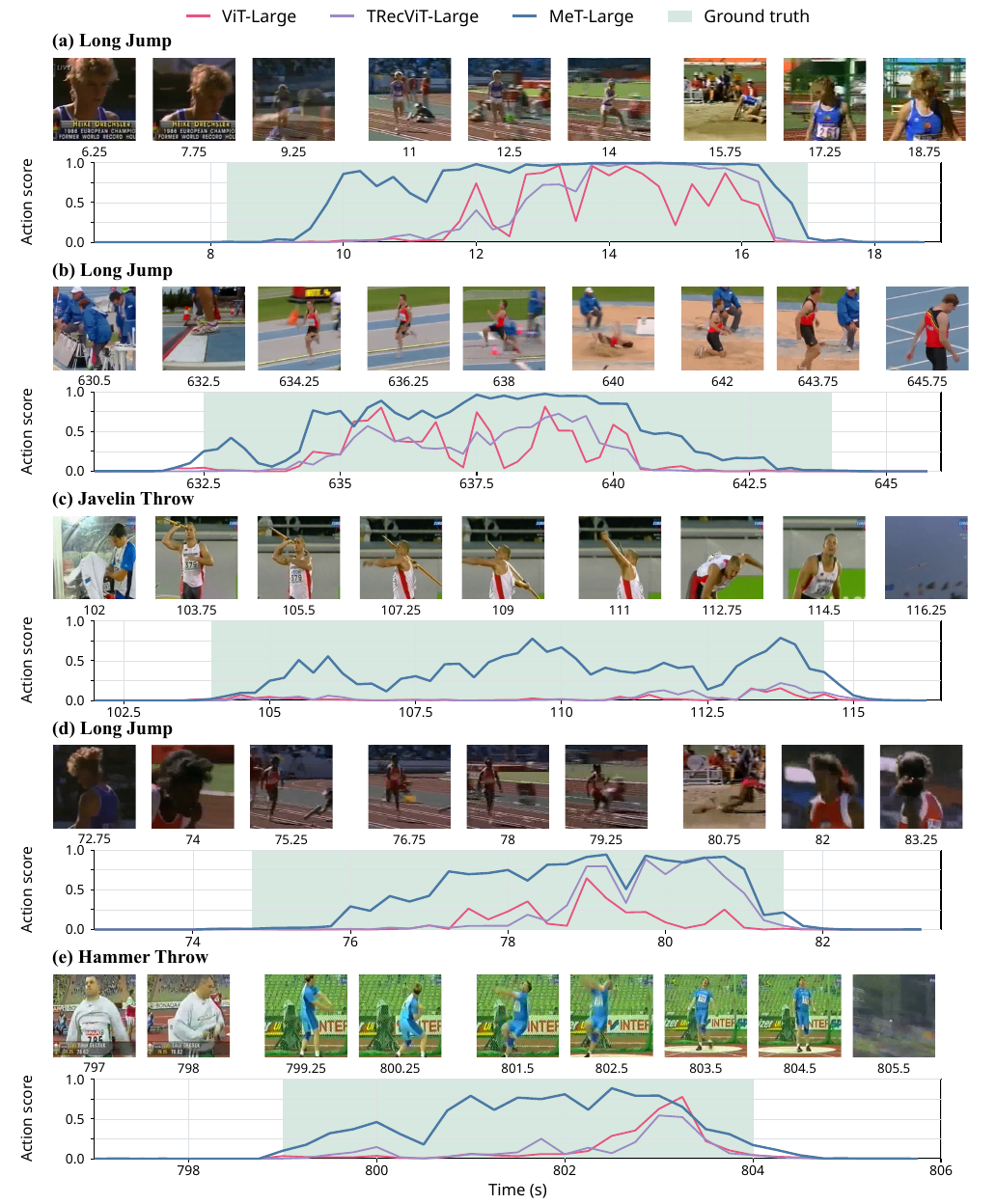}
    \caption{\textbf{Additional examples of causal framewise action recognition on THUMOS14.} Green shading marks the ground-truth action intervals. MeT-Large sustains higher scores than both baselines across much of each event, including the build-up to jumps and throws.}
    \label{fig:supp:thumos_qualitative}
\end{figure}

On THUMOS14~\cite{jiang2014thumos}, MeT-Large's \emph{long jump} score in Fig.~\ref{fig:supp:thumos_qualitative}(a) rises early in the run-up, ahead of both baselines, and stays high through most of the action. TRecViT-Large approaches that level around its later peaks. In Fig.~\ref{fig:supp:thumos_qualitative}(c), MeT-Large sustains a response through much of the \emph{javelin throw}, where both baselines stay near zero.

\begin{figure}[!p]
    \centering
    \includegraphics[width=\textwidth]{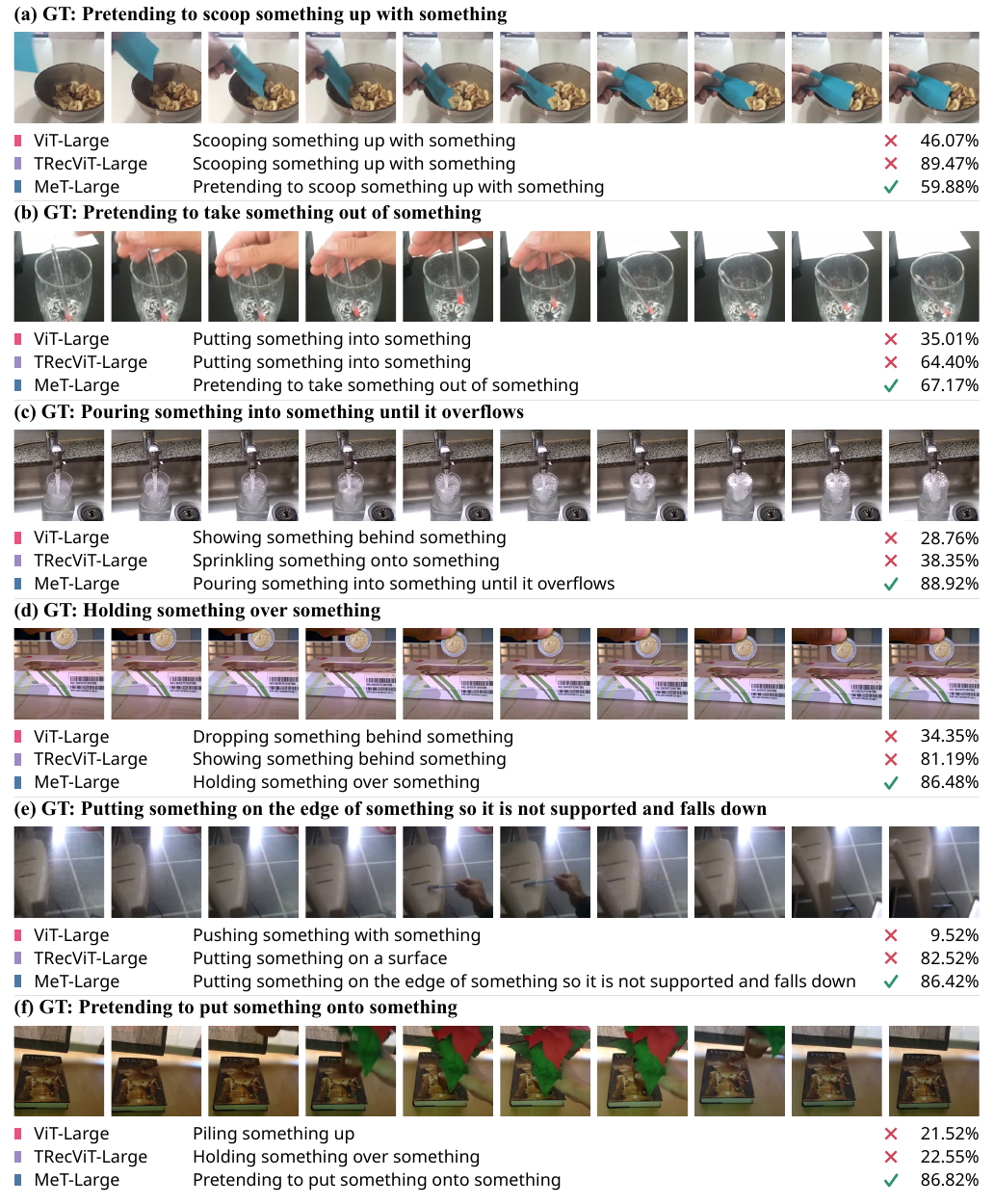}
    \caption{\textbf{Additional examples of clip-level action classification on SSv2.} Percentages are the softmax scores of the predicted classes. MeT-Large correctly identifies actions distinguished by object relations, mimed motions, or interaction outcomes.}
    \label{fig:supp:ssv2_qualitative}
\end{figure}

On SSv2~\cite{goyal2017something}, MeT-Large recognizes an object held \emph{over} another in Fig.~\ref{fig:supp:ssv2_qualitative}(d), where both baselines predict \emph{behind}. In Fig.~\ref{fig:supp:ssv2_qualitative}(e), it identifies the object's fall after placement on an edge, which distinguishes the action from simply placing the object on a surface.

\onecolumn
\captionsetup[figure]{hypcap=true}

\clearpage

\end{document}

%% file: preamble.tex
\usepackage{graphicx}
\usepackage{caption}

\usepackage{amsmath,amssymb}
\input{table_support}
\usepackage{placeins}
\input{figures/figure_04_caption}
\input{figures/figure_05_caption}

\input{figures/figure_01_caption}
\input{figures/figure_06_caption}
\usepackage{needspace}

%% file: table_support.tex
\usepackage{booktabs,array,multirow}
\usepackage{colortbl}
\usepackage{arydshln}
\usepackage{pifont}
\usepackage{caption}
\newcommand{\metyes}{\textcolor{TemporalYes}{\ding{51}}}
\newcommand{\metno}{\textcolor{TemporalNo}{\ding{55}}}
\newcommand{\metup}[1]{\,\textcolor{darkpastelgreen}{$\scriptstyle\blacktriangle\,#1$}}
\newcommand{\metdown}[1]{\,\textcolor{darkpink}{$\scriptstyle\blacktriangledown\,#1$}}
\input{tables/table_palette}

\input{tables/table_layout}
\newcommand{\metresourceup}[1]{\,\textcolor{ResourceDelta}{$\scriptstyle\blacktriangle\,#1$}}

\newsavebox{\mettablebox}
\makeatletter
\newcommand{\metdash}{\raisebox{-\dp\@arstrutbox}{%
  \vbox to \dimexpr\ht\@arstrutbox+\dp\@arstrutbox\relax{%
    \offinterlineskip\cleaders\vbox{%
      \hrule width\arrayrulewidth height1.2pt\kern1.4pt}\vfill}}}
\makeatother
\newcommand{\mettablestyle}{\setlength{\tabcolsep}{6pt}\renewcommand{\arraystretch}{1.20}}
\newcommand{\metcomparisonup}[1]{\,\textcolor{ComparisonDelta}{$\scriptstyle\blacktriangle\,#1$}}
\newcommand{\metcomparisondown}[1]{\,\textcolor{ComparisonDelta}{$\scriptstyle\blacktriangledown\,#1$}}

%% file: tables/table_palette.tex
\definecolor{AblationGreen}{RGB}{193,220,206}
\definecolor{darkpink}{rgb}{0.91,0.33,0.5}
\definecolor{darkpastelgreen}{rgb}{0.01,0.75,0.24}
\definecolor{ResourceBase}{RGB}{208,221,229}
\definecolor{ResourceDelta}{RGB}{74,108,130}
\definecolor{TemporalYes}{RGB}{66,125,91}
\definecolor{TemporalNo}{RGB}{175,85,103}
\definecolor{ComparisonDelta}{RGB}{164,92,52}

%% file: tables/table_layout.tex
\newcolumntype{C}[1]{>{\centering\arraybackslash}m{#1}}
\newlength{\YesWidth}
\newlength{\CausalWidth}
\newlength{\TemporalGap}
\newlength{\TypeWidth}
\newlength{\AccWidth}
\newlength{\ParamWidth}
\newlength{\StateWidth}
\newlength{\FlopsWidth}
\newlength{\LatencyWidth}
\newlength{\PeakWidth}
\newcommand{\HeaderOne}[1]{\begin{tabular}[c]{@{}c@{}}\bfseries #1\end{tabular}}
\newcommand{\HeaderLeft}[1]{\begin{tabular}[c]{@{}l@{}}\bfseries #1\end{tabular}}
\newcommand{\HeaderTemporal}{%
  \begingroup\renewcommand{\arraystretch}{1.0}%
  \begin{tabular}[c]{@{}C{\YesWidth}@{\hspace{\TemporalGap}}C{\CausalWidth}@{\hspace{\TemporalGap}}C{\TypeWidth}@{}}%
  \multicolumn{3}{c}{\bfseries Temporal Modeling}\\[2pt]
  \textcolor{black}{\pmb{\ding{51}}\,/\,\pmb{\ding{55}}} & \shortstack[c]{\bfseries Is\\\bfseries Causal} & \bfseries Type
  \end{tabular}\endgroup}
\newcommand{\HeaderAccuracy}[1]{%
  \begingroup\renewcommand{\arraystretch}{1.0}%
  \begin{tabular}[c]{@{}c@{}}\bfseries THUMOS14\\[2pt]
  \bfseries\boldmath #1\end{tabular}\endgroup}
\newcommand{\HeaderSSAccuracy}{%
  \begingroup\renewcommand{\arraystretch}{1.0}%
  \begin{tabular}[c]{@{}C{\AccWidth}C{\AccWidth}@{}}\multicolumn{2}{c}{\bfseries SSv2}\\[2pt]
  \bfseries\boldmath Top-1 $\uparrow$ & \bfseries\boldmath Top-5 $\uparrow$
  \end{tabular}\endgroup}
\newcommand{\HeaderResources}{%
  \begingroup\renewcommand{\arraystretch}{1.0}%
  \begin{tabular}[c]{@{}C{\ParamWidth}C{\StateWidth}C{\FlopsWidth}C{\LatencyWidth}C{\PeakWidth}@{}}
  \multicolumn{5}{c}{\bfseries Capacity \& Inference Efficiency}\\[2pt]
  \mbox{\bfseries\boldmath Param $\downarrow$} & \mbox{\bfseries\boldmath State/stream $\downarrow$} &
  \mbox{\bfseries\boldmath GFLOPs $\downarrow$} & \mbox{\bfseries\boldmath Latency $\downarrow$} & \mbox{\bfseries\boldmath GPU peak $\downarrow$}\\
  \bfseries (M) & \bfseries (MB) & \bfseries /frame & \bfseries (ms/frame) & \bfseries (GB)
  \end{tabular}\endgroup}
\newsavebox{\TypeBox}

\newsavebox{\TypeMeasure}
\newcommand{\TypeFit}[1]{%
  \sbox{\TypeMeasure}{#1}%
  \ifdim\wd\TypeMeasure>\TypeWidth
    \sbox{\TypeBox}{\parbox[c]{\TypeWidth}{\centering\hyphenpenalty=10000 #1\par}}%
    \raisebox{0pt}[\dimexpr\ht\TypeBox+3pt\relax][\dimexpr\dp\TypeBox+3pt\relax]{\usebox{\TypeBox}}%
  \else
    \raisebox{0pt}[\dimexpr\ht\TypeMeasure+1.5pt\relax][\dimexpr\dp\TypeMeasure+1.5pt\relax]{\usebox{\TypeMeasure}}%
  \fi}

%% file: figures/figure_04_caption.tex
\newcommand{\FigureFourSlantedPlusAux}[2]{\begingroup\setbox0=\hbox{$#1+$}\setbox2=\hbox{\pdfsave\pdfsetmatrix{1 0 0.29304564 1}\rlap{\copy0}\pdfrestore}\ht2=\ht0\dp2=\dp0\hbox to\wd0{\kern-.088\wd0\box2\kern1.088\wd0\hss}\endgroup}
\newcommand{\FigureFourItalicPlus}{\mathbin{\mathpalette\FigureFourSlantedPlusAux{}}}
\newcommand{\FigureFourCaption}{\textbf{The Memory Transformer (MeT) architecture.} \textit{Left:} Each MeT Block combines a Standard ViT Block with Token Memory. Patch+CLS Token Memory updates all tokens; CLS Token Memory uses the same PreLN-GRU structure for CLS alone. \textit{Right:} The panels detail the four designs compared in Tab.~\ref{tab:03_mechanism}; PreLN-GRU works best. $D$ is the spatial feature width, $H$ the GRU hidden width, and $\ell$ the spatial-block index. $h_t$, $S_t$, and $C_t$ denote recurrent hidden states, slots, and convolution caches at time $t$. In Slot Memory, \textit{Read} denotes current patch-and-CLS tokens, \textit{Patch} the original patches, and \textit{History} the preceding slots; $Q\FigureFourItalicPlus S$ combines learned queries $Q$ with those slots for writing. LN denotes LayerNorm.}

%% file: figures/figure_05_caption.tex
\newcommand{\FigureFiveCaption}{\textbf{Token-level state updates in MeT.} At intermediate depth $\ell$, CLS and $n$ fixed patch positions retain distinct states across frames. Inputs are normalized backbone features that incorporate spatial interactions. GRU parameters $\theta_\ell$ are shared across positions and time; different depths use separate modules and states. MeT retains token histories with shared parameters.}

%% file: figures/figure_01_caption.tex
\newcommand{\metcapacitycaption}{\textbf{Scaling memory capacity and model size consistently improves MeT accuracy, with compact states and low computational overhead relative to same-size ViTs.} Within each row, the spatial backbone and memory layout are fixed; ticks give the GRU hidden width $H$. Each size uses the same ImageNet-1K initialization across capacities; MeTs and ViTs follow the same training and evaluation protocols. Accuracy results are averaged over three random seeds. Pink dashed lines mark the corresponding ViT; green arrows show accuracy gains in percentage points. The five resource columns use the THUMOS14 configuration and share a vertical scale within each column. State/stream, Latency and GPU peak follow Fig.~\ref{fig:intro_comparison}. Native memory expands the scaling space of vision Transformers beyond backbone size, delivering sustained accuracy gains at nearly constant per-frame latency within each model size.}

%% file: figures/figure_06_caption.tex
\newcommand{\QualitativeCaption}{\textbf{MeT resolves visually confusable actions.} \textit{Left (THUMOS14):} Framewise scores for causal online action recognition; green shading marks the ground-truth (GT) event. \textit{Right (SSv2):} Predictions for two action sequences; percentages are softmax scores of the predicted classes. MeT translates evolving visual evidence into discriminative semantic judgments, supporting visual understanding at both frame and sequence levels.}

%% file: tables/table_03_mechanism.tex
\captionof{table}{\textbf{Comparison of single-location memory designs on a ViT-Small backbone.} Memory designs are compared at a common mid-depth position (block 6 of 12), using patch and CLS tokens under the same task recipe. Columns report THUMOS14 mAP and SSv2 Top-1; pink triangles show deficits relative to PreLN-GRU. PreLN-GRU leads both tasks, outperforming the best alternative in each by 1.30 mAP and 1.69 Top-1 points.}
\label{tab:03_mechanism}
\centering
\begingroup\mettablestyle
\ADLinactivate
\sbox{\mettablebox}{%
\begin{tabular}{l|c!{\metdash}c}
\toprule
{\bfseries\boldmath Memory Design} & {\bfseries\boldmath mAP $\uparrow$} & {\bfseries\boldmath Top-1 $\uparrow$} \\
\midrule
ViT-Small~\cite{dosovitskiy2021vit} & 38.94\metdown{3.13} & 30.59\metdown{21.47} \\
\midrule
\midrule
RG-LRU & 40.72\metdown{1.35} & 50.37\metdown{1.69} \\
Window Attention & 40.77\metdown{1.30} & 49.65\metdown{2.41} \\
Slot Memory & 39.86\metdown{2.21} & 49.42\metdown{2.64} \\
\multicolumn{1}{>{\columncolor{AblationGreen}[\tabcolsep][\tabcolsep]}l|}{PreLN-GRU} & \multicolumn{1}{>{\columncolor{AblationGreen}[\dimexpr\tabcolsep-\arrayrulewidth\relax][\dimexpr2\tabcolsep+\arrayrulewidth\relax]}c!{\textcolor{black}{\metdash}}}{42.07} & \multicolumn{1}{>{\columncolor{AblationGreen}[0pt][\tabcolsep]}c}{52.06} \\
\bottomrule
\end{tabular}}%
\typeout{MET-NATURAL-WIDTH: \the\wd\mettablebox; SLOT: \the\linewidth}
\scalebox{0.877921874}{\usebox{\mettablebox}}%
\endgroup

%% file: tables/table_04_content_position.tex
\captionof{table}{\textbf{Effects of memory inputs and insertion depth in ViT-Small.} L$x$ denotes memory insertion after spatial block $x$. Columns report THUMOS14 mAP and SSv2 Top-1; triangles show differences from L6: Patch + CLS. Joint patch-and-CLS memory at mid-depth improves both tasks over either input alone and avoids the \mbox{0.80--5.75}-point Top-1 losses of deeper insertion.}
\label{tab:04_content_position}
\centering
\begingroup\mettablestyle
\ADLinactivate
\sbox{\mettablebox}{%
\begin{tabular}{l|c!{\metdash}c}
\toprule
{\bfseries\boldmath Memory Input} & {\bfseries\boldmath mAP $\uparrow$} & {\bfseries\boldmath Top-1 $\uparrow$} \\
\midrule
\multicolumn{1}{>{\columncolor{AblationGreen}[\tabcolsep][\tabcolsep]}l|}{L6: Patch + CLS} & \multicolumn{1}{>{\columncolor{AblationGreen}[\dimexpr\tabcolsep-\arrayrulewidth\relax][\dimexpr2\tabcolsep+\arrayrulewidth\relax]}c!{\textcolor{black}{\metdash}}}{42.07} & \multicolumn{1}{>{\columncolor{AblationGreen}[0pt][\tabcolsep]}c}{52.06} \\
L6: Patch & 41.57\metdown{0.50} & 51.33\metdown{0.73} \\
L6: CLS & 40.92\metdown{1.15} & 47.28\metdown{4.78} \\
\midrule
\midrule
L3: Patch + CLS & 41.11\metdown{0.96} & 50.92\metdown{1.14} \\
L9: Patch + CLS & 42.83\metup{0.76} & 51.26\metdown{0.80} \\
L12: Patch + CLS & 42.92\metup{0.85} & 46.31\metdown{5.75} \\
\bottomrule
\end{tabular}}%
\typeout{MET-NATURAL-WIDTH: \the\wd\mettablebox; SLOT: \the\linewidth}
\scalebox{0.877921874}{\usebox{\mettablebox}}%
\endgroup

%% file: tables/table_05_layout.tex
\captionof{table}{\textbf{Comparison of distributed memory layouts for MeT-Small.} With no further spatial mixing before CLS readout, memory at block 12 processes CLS only; earlier modules process patch and CLS tokens. Triangles show differences from the highlighted layout. Resource columns use the THUMOS14 configuration. State/stream follows Fig.~\ref{fig:intro_comparison}. The four-stage hierarchy (\mbox{L3+L6+L9+L12}) achieves the highest THUMOS14 mAP and stays within 0.41 Top-1 points of all-block memory on SSv2, with 72.7\% less state.}
\label{tab:05_layout}
\centering
\begingroup\mettablestyle
\ADLinactivate
\newcommand{\metcostup}[1]{\,\textcolor{darkpink}{$\scriptstyle\blacktriangle\,#1$}}
\newcommand{\metcostdown}[1]{\,\textcolor{darkpastelgreen}{$\scriptstyle\blacktriangledown\,#1$}}

\newsavebox{\metlayoutnamebox}
\sbox{\metlayoutnamebox}{\begin{tabular}[c]{@{}l@{}}%
L1+L3+L4+L6\\
+L7+L9+L10+L12
\end{tabular}}
\newcommand{\metlayoutlongdash}{\raisebox{-\dp\metlayoutnamebox}{%
  \vbox to \dimexpr\ht\metlayoutnamebox+\dp\metlayoutnamebox\relax{%
    \offinterlineskip\cleaders\vbox{%
      \hrule width\arrayrulewidth height1.2pt\kern1.4pt}\vfill}}}
\typeout{MET-LAYOUT-LONGROW: HEIGHT \the\ht\metlayoutnamebox; DEPTH \the\dp\metlayoutnamebox}
\sbox{\mettablebox}{%
\begin{tabular}{l|c|c!{\metdash}c|cc}
\toprule
{\bfseries\boldmath After Spatial Blocks} & {\bfseries\boldmath Token Policy} & {\bfseries\boldmath THUMOS14 mAP $\uparrow$} & {\bfseries\boldmath SSv2 Top-1 $\uparrow$} & {\bfseries\boldmath Param (M) $\downarrow$} & {\bfseries\boldmath State/stream (MB) $\downarrow$} \\
\midrule
L6 & All: Patch+CLS & 42.07\metdown{2.86} & 52.06\metdown{3.60} & 22.71\metcostdown{3.11} & 0.30\metcostdown{0.61} \\
\midrule
L3+L6 & All: Patch+CLS & 40.74\metdown{4.19} & 53.96\metdown{1.70} & 23.75\metcostdown{2.07} & 0.61\metcostdown{0.30} \\
L6+L9 & All: Patch+CLS & 43.18\metdown{1.75} & 53.99\metdown{1.67} & 23.75\metcostdown{2.07} & 0.61\metcostdown{0.30} \\
\midrule
L3+L6+L9 & All: Patch+CLS & 42.73\metdown{2.20} & 54.89\metdown{0.77} & 24.78\metcostdown{1.04} & 0.908\metcostdown{0.002} \\
L6+L9+L12 & L12: CLS, Other: Patch+CLS & 44.26\metdown{0.67} & 54.74\metdown{0.92} & 24.78\metcostdown{1.04} & 0.61\metcostdown{0.30} \\
\midrule
L1+L4+L7+L10 & All: Patch+CLS & 41.45\metdown{3.48} & 56.22\metup{0.56} & 25.82 & 1.21\metcostup{0.30} \\
L2+L5+L8+L11 & All: Patch+CLS & 43.68\metdown{1.25} & 56.23\metup{0.57} & 25.82 & 1.21\metcostup{0.30} \\
\multicolumn{1}{>{\columncolor{AblationGreen}[\tabcolsep][\tabcolsep]}l|}{L3+L6+L9+L12} & \multicolumn{1}{>{\columncolor{AblationGreen}[\dimexpr\tabcolsep-\arrayrulewidth\relax][\tabcolsep]}c|}{L12: CLS, Other: Patch+CLS} & \multicolumn{1}{>{\columncolor{AblationGreen}[\dimexpr\tabcolsep-\arrayrulewidth\relax][\dimexpr2\tabcolsep+\arrayrulewidth\relax]}c!{\textcolor{black}{\metdash}}}{44.93} & \multicolumn{1}{>{\columncolor{AblationGreen}[0pt][\tabcolsep]}c|}{55.66} & \multicolumn{1}{>{\columncolor{AblationGreen}[\dimexpr\tabcolsep-\arrayrulewidth\relax][\tabcolsep]}c}{25.82} & \multicolumn{1}{>{\columncolor{AblationGreen}[\tabcolsep][\tabcolsep]}c}{0.91} \\
\midrule
\usebox{\metlayoutnamebox} & \multicolumn{1}{c|}{L12: CLS, Other: Patch+CLS} & \multicolumn{1}{c!{\textcolor{black}{\metlayoutlongdash}}}{44.36\metdown{0.57}} & 55.94\metup{0.28} & 29.96\metcostup{4.14} & 2.12\metcostup{1.21} \\
\midrule
All 12 Blocks & L12: CLS, Other: Patch+CLS & 42.92\metdown{2.01} & 56.07\metup{0.41} & 34.11\metcostup{8.29} & 3.33\metcostup{2.42} \\
\bottomrule
\end{tabular}}%
\typeout{MET-NATURAL-WIDTH: \the\wd\mettablebox; SLOT: \the\linewidth}
\ifdim\wd\mettablebox>\linewidth\resizebox{\linewidth}{!}{\usebox{\mettablebox}}%
\else\usebox{\mettablebox}\fi
\endgroup

%% file: tables/table_06_family.tex
\captionof{table}{\textbf{Details of MeT models.} By default, we match the GRU hidden width $H$ to ViT's hidden width and retain its original number of Transformer blocks $N$. Parameter and state counts are for \mbox{THUMOS14}; State/stream follows Fig.~\ref{fig:intro_comparison}.}
\label{tab:06_family}
\centering
\begingroup\mettablestyle
\setlength{\tabcolsep}{3.5pt}
\ADLinactivate
\newcommand{\metconfigcell}[1]{\begin{tabular}[c]{@{}c@{}}#1\end{tabular}}
\newsavebox{\metconfigheadbox}
\sbox{\metconfigheadbox}{{\bfseries\metconfigcell{State/stream\\(MB)}}}
\newcommand{\metconfigheaddash}{\raisebox{-\dp\metconfigheadbox}{%
  \vbox to \dimexpr\ht\metconfigheadbox+\dp\metconfigheadbox\relax{%
    \offinterlineskip\cleaders\vbox{%
      \hrule width\arrayrulewidth height1.2pt\kern1.4pt}\vfill}}}

\sbox{\mettablebox}{%
\begin{tabular}{l|cc!{\metdash}c|cc}
\toprule
\textbf{Model} & {\bfseries\boldmath $H$} & \multicolumn{1}{c!{\metconfigheaddash}}{{\bfseries\boldmath $N$}} & \textbf{Memory after Blocks} & {\bfseries\metconfigcell{Param\\(M)}} & {\bfseries\metconfigcell{State/stream\\(MB)}} \\
\midrule
MeT-Small & 384 & 12 & 3, 6, 9, 12 & 25.82 & 0.91 \\
MeT-Base & 768 & 12 & 3, 6, 9, 12 & 102.36 & 1.82 \\
MeT-Large & 1024 & 24 & 3, 6, 9, 12, 15, 18, 21, 24 & 362.13 & 5.65 \\
\bottomrule
\end{tabular}}%
\typeout{MET-NATURAL-WIDTH: \the\wd\mettablebox; SLOT: \the\linewidth}
\ifdim\wd\mettablebox>\linewidth
\resizebox{\linewidth}{!}{\usebox{\mettablebox}}%
\else\usebox{\mettablebox}\fi
\endgroup

%% file: tables/table_01_thumos.tex
\captionof{table}{\textbf{Comparison with representative causal action recognition methods on THUMOS14~\cite{jiang2014thumos}.} Pink triangles show mAP differences from MeT-Large; resource differences compare MeT with the same-size ViT (blue triangles) and other methods with ViT-Large (copper triangles). Is Causal indicates whether framewise computation uses only current and past inputs. State/stream, Latency and GPU peak follow Fig.~\ref{fig:intro_comparison}. V. Mamba Suite denotes Video Mamba Suite. MeT-Large achieves the highest mAP, surpassing TRecViT-Large by 2.28 points with 92.7\% less state and a $3.0\times$ causal framewise speedup.}
\label{tab:01_thumos}
\centering
\begingroup\mettablestyle
\ADLinactivate
\sbox{\mettablebox}{%
\begin{tabular}{l|c|C{\YesWidth}@{\hspace{\TemporalGap}}C{\CausalWidth}@{\hspace{\TemporalGap}}C{\TypeWidth}|c|C{\ParamWidth}C{\StateWidth}C{\FlopsWidth}C{\LatencyWidth}C{\PeakWidth}}
\toprule
\HeaderLeft{Method} & \HeaderOne{Pub \& Year} & \multicolumn{3}{c|}{\HeaderTemporal} & \HeaderAccuracy{mAP $\uparrow$} & \multicolumn{5}{c}{\HeaderResources} \\
\midrule
ViT-Small~\cite{dosovitskiy2021vit} & ICLR 2021 & \metno & \metyes & \textemdash & 38.94\metdown{10.28} & \multicolumn{1}{>{\columncolor{ResourceBase}[\dimexpr\tabcolsep-\arrayrulewidth\relax][\tabcolsep]}C{\ParamWidth}}{21.67} & \multicolumn{1}{>{\columncolor{ResourceBase}[\tabcolsep][\tabcolsep]}C{\StateWidth}}{0.00} & \multicolumn{1}{>{\columncolor{ResourceBase}[\tabcolsep][\tabcolsep]}C{\FlopsWidth}}{9.20} & \multicolumn{1}{>{\columncolor{ResourceBase}[\tabcolsep][\tabcolsep]}C{\LatencyWidth}}{2.18} & \multicolumn{1}{>{\columncolor{ResourceBase}[\tabcolsep][\tabcolsep]}C{\PeakWidth}}{0.48} \\
ViT-Base~\cite{dosovitskiy2021vit} & ICLR 2021 & \metno & \metyes & \textemdash & 40.67\metdown{8.55} & \multicolumn{1}{>{\columncolor{ResourceBase}[\dimexpr\tabcolsep-\arrayrulewidth\relax][\tabcolsep]}C{\ParamWidth}}{85.81} & \multicolumn{1}{>{\columncolor{ResourceBase}[\tabcolsep][\tabcolsep]}C{\StateWidth}}{0.00} & \multicolumn{1}{>{\columncolor{ResourceBase}[\tabcolsep][\tabcolsep]}C{\FlopsWidth}}{35.13} & \multicolumn{1}{>{\columncolor{ResourceBase}[\tabcolsep][\tabcolsep]}C{\LatencyWidth}}{2.21} & \multicolumn{1}{>{\columncolor{ResourceBase}[\tabcolsep][\tabcolsep]}C{\PeakWidth}}{1.05} \\
ViT-Large~\cite{dosovitskiy2021vit} & ICLR 2021 & \metno & \metyes & \textemdash & 42.53\metdown{6.69} & \multicolumn{1}{>{\columncolor{ResourceBase}[\dimexpr\tabcolsep-\arrayrulewidth\relax][\tabcolsep]}C{\ParamWidth}}{303.32} & \multicolumn{1}{>{\columncolor{ResourceBase}[\tabcolsep][\tabcolsep]}C{\StateWidth}}{0.00} & \multicolumn{1}{>{\columncolor{ResourceBase}[\tabcolsep][\tabcolsep]}C{\FlopsWidth}}{123.11} & \multicolumn{1}{>{\columncolor{ResourceBase}[\tabcolsep][\tabcolsep]}C{\LatencyWidth}}{4.04} & \multicolumn{1}{>{\columncolor{ResourceBase}[\tabcolsep][\tabcolsep]}C{\PeakWidth}}{2.13} \\
\midrule
TRecViT-Large~\cite{patraucean2026trecvit} & TMLR 2026 & \metyes & \metyes & \TypeFit{Tokenwise recurrence} & 46.94\metdown{2.28} & 382.26\metcomparisonup{78.94} & 77.46\metcomparisonup{77.46} & 154.09\metcomparisonup{30.98} & 18.46\metcomparisonup{14.42} & 3.59\metcomparisonup{1.46} \\
ViT-Large + CMeRT~\cite{pang2025cmert} & CVPR 2025 & \metyes & \metyes & \TypeFit{Memory refinement} & 45.85\metdown{3.37} & 389.58\metcomparisonup{86.26} & 4.27\metcomparisonup{4.27} & 129.79\metcomparisonup{6.68} & 8.56\metcomparisonup{4.52} & 2.50\metcomparisonup{0.37} \\
ViT-Large + MiniROAD~\cite{an2023miniroad} & ICCV 2023 & \metyes & \metyes & \TypeFit{Recurrent head} & 44.88\metdown{4.34} & 314.87\metcomparisonup{11.55} & 0.004\metcomparisonup{0.004} & 123.13\metcomparisonup{0.02} & 4.08\metcomparisonup{0.04} & 2.18\metcomparisonup{0.05} \\
ViT-Large + MAT~\cite{wang2023memory} & ICCV 2023 & \metyes & \metyes & \TypeFit{Memory + anticipation} & 45.68\metdown{3.54} & 402.16\metcomparisonup{98.84} & 4.33\metcomparisonup{4.33} & 133.38\metcomparisonup{10.27} & 11.82\metcomparisonup{7.78} & 2.55\metcomparisonup{0.42} \\
ViT-Large + TeSTra~\cite{zhao2022testra} & ECCV 2022 & \metyes & \metyes & \TypeFit{Temporal smoothing} & 44.72\metdown{4.50} & 357.99\metcomparisonup{54.67} & 8.52\metcomparisonup{8.52} & 129.62\metcomparisonup{6.51} & 6.13\metcomparisonup{2.09} & 2.37\metcomparisonup{0.24} \\
ViT-Large + Colar~\cite{yang2022colar} & CVPR 2022 & \metyes & \metyes & \TypeFit{Exemplar consultation} & 45.40\metdown{3.82} & 319.11\metcomparisonup{15.79} & 0.26\metcomparisonup{0.26} & 125.07\metcomparisonup{1.96} & 8.01\metcomparisonup{3.97} & 2.19\metcomparisonup{0.06} \\
ViT-Large + V. Mamba Suite~\cite{chen2026videomambasuite} & IJCV 2026 & \metyes & \metyes & \TypeFit{State-space head} & 44.42\metdown{4.80} & 317.71\metcomparisonup{14.39} & 0.13\metcomparisonup{0.13} & 124.02\metcomparisonup{0.91} & 7.16\metcomparisonup{3.12} & 2.21\metcomparisonup{0.08} \\
\midrule
MeT-Small & Ours & \metyes & \metyes & \TypeFit{Hierarchical native memory} & 44.93\metdown{4.29} & 25.82\metresourceup{4.15} & 0.91\metresourceup{0.91} & 10.42\metresourceup{1.22} & 3.19\metresourceup{1.01} & 0.51\metresourceup{0.03} \\
MeT-Base & Ours & \metyes & \metyes & \TypeFit{Hierarchical native memory} & 47.96\metdown{1.26} & 102.36\metresourceup{16.55} & 1.82\metresourceup{1.82} & 40.01\metresourceup{4.88} & 3.20\metresourceup{0.99} & 1.13\metresourceup{0.08} \\
\multicolumn{1}{>{\columncolor{AblationGreen}[\tabcolsep][\tabcolsep]}l|}{MeT-Large} & \multicolumn{1}{>{\columncolor{AblationGreen}[\dimexpr\tabcolsep-\arrayrulewidth\relax][\tabcolsep]}c|}{Ours} & \multicolumn{1}{>{\columncolor{AblationGreen}[\dimexpr\tabcolsep-\arrayrulewidth\relax][\tabcolsep]}C{\YesWidth}@{\hspace{\TemporalGap}}}{\metyes} & \multicolumn{1}{@{}>{\columncolor{AblationGreen}[\tabcolsep][\tabcolsep]}C{\CausalWidth}@{\hspace{\TemporalGap}}}{\metyes} & \multicolumn{1}{@{}>{\columncolor{AblationGreen}[\tabcolsep][\tabcolsep]}C{\TypeWidth}|}{\TypeFit{Hierarchical native memory}} & \multicolumn{1}{>{\columncolor{AblationGreen}[\dimexpr\tabcolsep-\arrayrulewidth\relax][\tabcolsep]}c|}{49.22} & \multicolumn{1}{>{\columncolor{AblationGreen}[\dimexpr\tabcolsep-\arrayrulewidth\relax][\tabcolsep]}C{\ParamWidth}}{362.13\metresourceup{58.81}} & \multicolumn{1}{>{\columncolor{AblationGreen}[\tabcolsep][\tabcolsep]}C{\StateWidth}}{5.66\metresourceup{5.66}} & \multicolumn{1}{>{\columncolor{AblationGreen}[\tabcolsep][\tabcolsep]}C{\FlopsWidth}}{143.37\metresourceup{20.26}} & \multicolumn{1}{>{\columncolor{AblationGreen}[\tabcolsep][\tabcolsep]}C{\LatencyWidth}}{6.18\metresourceup{2.14}} & \multicolumn{1}{>{\columncolor{AblationGreen}[\tabcolsep][\tabcolsep]}C{\PeakWidth}}{2.41\metresourceup{0.28}} \\
\bottomrule
\end{tabular}}%
\typeout{MET-NATURAL-WIDTH: \the\wd\mettablebox; SLOT: \the\linewidth}
\ifdim\wd\mettablebox>\linewidth\resizebox{\linewidth}{!}{\usebox{\mettablebox}}%
\else\usebox{\mettablebox}\fi
\endgroup

%% file: tables/table_02_ssv2.tex
\captionof{table}{\textbf{Comparison with representative video classification methods on SSv2~\cite{goyal2017something}.} Top-1 and Top-5 report full-clip classification from 32 sampled frames; Is Causal describes framewise computation. Pink triangles show accuracy differences from MeT-Large; other triangle colors and metric definitions follow Tab.~\ref{tab:01_thumos}. SSv2 state also counts the feature sum and frame counter used for clip averaging, including in ViT. Dashes in the State/stream and Latency columns indicate unsupported causal framewise inference. MeT-Small exceeds every non-MeT comparator, while MeT-Large widens the lead to 4.17 Top-1 and 6.85 Top-5 points.}
\label{tab:02_ssv2}
\centering
\begingroup\mettablestyle
\ADLinactivate
\sbox{\mettablebox}{%
\begin{tabular}{l|c|C{\YesWidth}@{\hspace{\TemporalGap}}C{\CausalWidth}@{\hspace{\TemporalGap}}C{\TypeWidth}|C{\AccWidth}C{\AccWidth}|C{\ParamWidth}C{\StateWidth}C{\FlopsWidth}C{\LatencyWidth}C{\PeakWidth}}
\toprule
\HeaderLeft{Method} & \HeaderOne{Pub \& Year} & \multicolumn{3}{c|}{\HeaderTemporal} & \multicolumn{2}{c|}{\HeaderSSAccuracy} & \multicolumn{5}{c}{\HeaderResources} \\
\midrule
ViT-Small~\cite{dosovitskiy2021vit} & ICLR 2021 & \metno & \metyes & \textemdash & 30.59\metdown{27.89} & 61.09\metdown{23.68} & \multicolumn{1}{>{\columncolor{ResourceBase}[\dimexpr\tabcolsep-\arrayrulewidth\relax][\tabcolsep]}C{\ParamWidth}}{21.73} & \multicolumn{1}{>{\columncolor{ResourceBase}[\tabcolsep][\tabcolsep]}C{\StateWidth}}{0.002} & \multicolumn{1}{>{\columncolor{ResourceBase}[\tabcolsep][\tabcolsep]}C{\FlopsWidth}}{9.20} & \multicolumn{1}{>{\columncolor{ResourceBase}[\tabcolsep][\tabcolsep]}C{\LatencyWidth}}{2.21} & \multicolumn{1}{>{\columncolor{ResourceBase}[\tabcolsep][\tabcolsep]}C{\PeakWidth}}{0.48} \\
ViT-Base~\cite{dosovitskiy2021vit} & ICLR 2021 & \metno & \metyes & \textemdash & 31.39\metdown{27.09} & 61.53\metdown{23.24} & \multicolumn{1}{>{\columncolor{ResourceBase}[\dimexpr\tabcolsep-\arrayrulewidth\relax][\tabcolsep]}C{\ParamWidth}}{85.93} & \multicolumn{1}{>{\columncolor{ResourceBase}[\tabcolsep][\tabcolsep]}C{\StateWidth}}{0.003} & \multicolumn{1}{>{\columncolor{ResourceBase}[\tabcolsep][\tabcolsep]}C{\FlopsWidth}}{35.13} & \multicolumn{1}{>{\columncolor{ResourceBase}[\tabcolsep][\tabcolsep]}C{\LatencyWidth}}{2.22} & \multicolumn{1}{>{\columncolor{ResourceBase}[\tabcolsep][\tabcolsep]}C{\PeakWidth}}{1.05} \\
ViT-Large~\cite{dosovitskiy2021vit} & ICLR 2021 & \metno & \metyes & \textemdash & 31.57\metdown{26.91} & 63.01\metdown{21.76} & \multicolumn{1}{>{\columncolor{ResourceBase}[\dimexpr\tabcolsep-\arrayrulewidth\relax][\tabcolsep]}C{\ParamWidth}}{303.48} & \multicolumn{1}{>{\columncolor{ResourceBase}[\tabcolsep][\tabcolsep]}C{\StateWidth}}{0.004} & \multicolumn{1}{>{\columncolor{ResourceBase}[\tabcolsep][\tabcolsep]}C{\FlopsWidth}}{123.11} & \multicolumn{1}{>{\columncolor{ResourceBase}[\tabcolsep][\tabcolsep]}C{\LatencyWidth}}{4.11} & \multicolumn{1}{>{\columncolor{ResourceBase}[\tabcolsep][\tabcolsep]}C{\PeakWidth}}{2.13} \\
\midrule
TRecViT-Large~\cite{patraucean2026trecvit} & TMLR 2026 & \metyes & \metyes & \TypeFit{Tokenwise recurrence} & 54.31\metdown{4.17} & 77.92\metdown{6.85} & 382.42\metcomparisonup{78.94} & 77.47\metcomparisonup{77.47} & 154.09\metcomparisonup{30.98} & 18.67\metcomparisonup{14.56} & 3.59\metcomparisonup{1.46} \\
VideoMamba-M~\cite{li2024videomamba} & ECCV 2024 & \metyes & \metno & \TypeFit{Bidirectional state space} & 51.88\metdown{6.60} & 77.44\metdown{7.33} & 73.76\metcomparisondown{229.72} & \textemdash & 29.92\metcomparisondown{93.19} & \textemdash & 1.30\metcomparisondown{0.83} \\
ViT-Large + Frame2Freq-MS~\cite{ponbagavathi2026frame2freq} & CVPR 2026 & \metyes & \metno & \TypeFit{Spectral adapters} & 43.56\metdown{14.92} & 67.80\metdown{16.97} & 322.44\metcomparisonup{18.96} & \textemdash & 130.56\metcomparisonup{7.45} & \textemdash & 2.11\metcomparisondown{0.02} \\
ViT-Large + DejaVid~\cite{ho2025dejavid} & CVPR 2025 & \metyes & \metno & \TypeFit{Temporal matching} & 39.41\metdown{19.07} & 65.73\metdown{19.04} & 316.82\metcomparisonup{13.34} & \textemdash & 123.16\metcomparisonup{0.05} & \textemdash & 2.24\metcomparisonup{0.11} \\
\midrule
MeT-Small & Ours & \metyes & \metyes & \TypeFit{Hierarchical native memory} & 55.66\metdown{2.82} & 79.54\metdown{5.23} & 25.88\metresourceup{4.15} & 0.91\metresourceup{0.91} & 10.42\metresourceup{1.22} & 3.24\metresourceup{1.03} & 0.51\metresourceup{0.03} \\
MeT-Base & Ours & \metyes & \metyes & \TypeFit{Hierarchical native memory} & 57.02\metdown{1.46} & 79.66\metdown{5.11} & 102.48\metresourceup{16.55} & 1.82\metresourceup{1.82} & 40.01\metresourceup{4.88} & 3.26\metresourceup{1.04} & 1.13\metresourceup{0.08} \\
\multicolumn{1}{>{\columncolor{AblationGreen}[\tabcolsep][\tabcolsep]}l|}{MeT-Large} & \multicolumn{1}{>{\columncolor{AblationGreen}[\dimexpr\tabcolsep-\arrayrulewidth\relax][\tabcolsep]}c|}{Ours} & \multicolumn{1}{>{\columncolor{AblationGreen}[\dimexpr\tabcolsep-\arrayrulewidth\relax][\tabcolsep]}C{\YesWidth}@{\hspace{\TemporalGap}}}{\metyes} & \multicolumn{1}{@{}>{\columncolor{AblationGreen}[\tabcolsep][\tabcolsep]}C{\CausalWidth}@{\hspace{\TemporalGap}}}{\metyes} & \multicolumn{1}{@{}>{\columncolor{AblationGreen}[\tabcolsep][\tabcolsep]}C{\TypeWidth}|}{\TypeFit{Hierarchical native memory}} & \multicolumn{1}{>{\columncolor{AblationGreen}[\dimexpr\tabcolsep-\arrayrulewidth\relax][\tabcolsep]}C{\AccWidth}}{58.48} & \multicolumn{1}{>{\columncolor{AblationGreen}[\tabcolsep][\tabcolsep]}C{\AccWidth}|}{84.77} & \multicolumn{1}{>{\columncolor{AblationGreen}[\dimexpr\tabcolsep-\arrayrulewidth\relax][\tabcolsep]}C{\ParamWidth}}{362.29\metresourceup{58.81}} & \multicolumn{1}{>{\columncolor{AblationGreen}[\tabcolsep][\tabcolsep]}C{\StateWidth}}{5.66\metresourceup{5.66}} & \multicolumn{1}{>{\columncolor{AblationGreen}[\tabcolsep][\tabcolsep]}C{\FlopsWidth}}{143.37\metresourceup{20.26}} & \multicolumn{1}{>{\columncolor{AblationGreen}[\tabcolsep][\tabcolsep]}C{\LatencyWidth}}{6.18\metresourceup{2.07}} & \multicolumn{1}{>{\columncolor{AblationGreen}[\tabcolsep][\tabcolsep]}C{\PeakWidth}}{2.41\metresourceup{0.28}} \\
\bottomrule
\end{tabular}}%
\typeout{MET-NATURAL-WIDTH: \the\wd\mettablebox; SLOT: \the\linewidth}
\ifdim\wd\mettablebox>\linewidth\resizebox{\linewidth}{!}{\usebox{\mettablebox}}%
\else\usebox{\mettablebox}\fi
\endgroup

%% file: tables/table_s01_size_comparison.tex
\captionof{table}{\textbf{Comparison of ViT, TRecViT, and MeT across backbone sizes.} Resource metrics use THUMOS14 configurations and follow the main paper's measurement protocol. Pink triangles show accuracy differences from the same-size MeT. Copper and blue triangles show resource differences for TRecViT and MeT relative to the same-size ViT, respectively. At each size, MeT achieves the highest mAP and Top-1 accuracy with 92.7--93.7\% less state per stream than TRecViT.}
\label{tab:supp:size_comparison}
\centering
\begingroup
\mettablestyle
\setlength{\tabcolsep}{2.5pt}
\small
\ADLinactivate
\settowidth{\AccWidth}{00.00\metdown{00.00}}
\newcommand{\suppGreenCell}[2]{\multicolumn{1}{>{\columncolor{AblationGreen}[\dimexpr\tabcolsep-\arrayrulewidth\relax][\tabcolsep]}c#1}{#2}}
\sbox{\mettablebox}{%
\begin{tabular}{l|c|c!{\metdash}C{\AccWidth}C{\AccWidth}|ccccc}
\toprule
\multirow{3}{*}{\textbf{Method}} & \multirow{3}{*}{\HeaderOne{Temporal\\\bfseries Modeling}} & \multirow{3}{*}{\HeaderAccuracy{mAP $\uparrow$}} & \multicolumn{2}{c|}{\multirow{3}{*}{\HeaderSSAccuracy}} & \multicolumn{5}{c}{\textbf{Capacity \& Inference Efficiency}} \\
 & & & & & \textbf{\boldmath Param $\downarrow$} & \textbf{\boldmath State/stream $\downarrow$} & \textbf{\boldmath GFLOPs $\downarrow$} & \textbf{\boldmath Latency $\downarrow$} & \textbf{\boldmath GPU peak $\downarrow$} \\
 & & & & & \textbf{(M)} & \textbf{(MB)} & \textbf{/frame} & \textbf{(ms/frame)} & \textbf{(GB)} \\
\midrule
ViT-Small~\cite{dosovitskiy2021vit} & \metno & 38.94\metdown{5.99} & 30.59\metdown{25.07} & 61.09\metdown{18.45} & \cellcolor{ResourceBase}21.67 & \cellcolor{ResourceBase}0.00 & \cellcolor{ResourceBase}9.20 & \cellcolor{ResourceBase}2.18 & \cellcolor{ResourceBase}0.48 \\
TRecViT-Small~\cite{patraucean2026trecvit} & \metyes & 42.92\metdown{2.01} & 52.05\metdown{3.61} & 77.52\metdown{2.02} & 27.63\metcomparisonup{5.96} & 14.52\metcomparisonup{14.52} & 11.52\metcomparisonup{2.32} & 8.79\metcomparisonup{6.61} & 0.82\metcomparisonup{0.34} \\
\rowcolor{AblationGreen} MeT-Small & \suppGreenCell{|}{\metyes} & \suppGreenCell{!{\metdash}}{44.93} & 55.66 & 79.54 & 25.82\metresourceup{4.15} & 0.91\metresourceup{0.91} & 10.42\metresourceup{1.22} & 3.19\metresourceup{1.01} & 0.51\metresourceup{0.03} \\
\midrule
ViT-Base~\cite{dosovitskiy2021vit} & \metno & 40.67\metdown{7.29} & 31.39\metdown{25.63} & 61.53\metdown{18.13} & \cellcolor{ResourceBase}85.81 & \cellcolor{ResourceBase}0.00 & \cellcolor{ResourceBase}35.13 & \cellcolor{ResourceBase}2.21 & \cellcolor{ResourceBase}1.05 \\
TRecViT-Base~\cite{patraucean2026trecvit} & \metyes & 45.95\metdown{2.01} & 53.47\metdown{3.55} & 77.63\metdown{2.03} & 108.34\metcomparisonup{22.53} & 29.05\metcomparisonup{29.05} & 43.96\metcomparisonup{8.83} & 8.80\metcomparisonup{6.59} & 1.77\metcomparisonup{0.72} \\
\rowcolor{AblationGreen} MeT-Base & \suppGreenCell{|}{\metyes} & \suppGreenCell{!{\metdash}}{47.96} & 57.02 & 79.66 & 102.36\metresourceup{16.55} & 1.82\metresourceup{1.82} & 40.01\metresourceup{4.88} & 3.20\metresourceup{0.99} & 1.13\metresourceup{0.08} \\
\midrule
ViT-Large~\cite{dosovitskiy2021vit} & \metno & 42.53\metdown{6.69} & 31.57\metdown{26.91} & 63.01\metdown{21.76} & \cellcolor{ResourceBase}303.32 & \cellcolor{ResourceBase}0.00 & \cellcolor{ResourceBase}123.11 & \cellcolor{ResourceBase}4.04 & \cellcolor{ResourceBase}2.13 \\
TRecViT-Large~\cite{patraucean2026trecvit} & \metyes & 46.94\metdown{2.28} & 54.31\metdown{4.17} & 77.92\metdown{6.85} & 382.26\metcomparisonup{78.94} & 77.46\metcomparisonup{77.46} & 154.09\metcomparisonup{30.98} & 18.46\metcomparisonup{14.42} & 3.59\metcomparisonup{1.46} \\
\rowcolor{AblationGreen} MeT-Large & \suppGreenCell{|}{\metyes} & \suppGreenCell{!{\metdash}}{49.22} & 58.48 & 84.77 & 362.13\metresourceup{58.81} & 5.66\metresourceup{5.66} & 143.37\metresourceup{20.26} & 6.18\metresourceup{2.14} & 2.41\metresourceup{0.28} \\
\bottomrule
\end{tabular}}%
\typeout{MET-SUPP-TABLE-WIDTH: \the\wd\mettablebox; SLOT: \the\linewidth}
\ifdim\wd\mettablebox>\linewidth
  \resizebox{\linewidth}{!}{\usebox{\mettablebox}}%
\else
  \usebox{\mettablebox}%
\fi
\par
\endgroup